\documentclass[titlepage]{article}
\usepackage[T1]{fontenc}
\usepackage[utf8]{inputenc}
\usepackage{subfigure}
\usepackage{color}
\usepackage[edges]{forest}
\usepackage{url}
\usepackage{microtype}
\usepackage{datetime}
\usepackage{authblk}
\usepackage{listings}
\usepackage{upquote}
\usepackage[symbol]{footmisc}

\usepackage{hyperref}

\newdateformat{monthyeardate}{\monthname[\THEMONTH] \THEYEAR}

\definecolor{foldercolor}{RGB}{124,166,198}

\forestset{is file/.style={edge path'/.expanded={%
        ([xshift=\forestregister{folder indent}]!u.parent anchor) |- (.child anchor)}, inner sep=-2pt},
    this folder size/.style={edge path'/.expanded={%
        ([xshift=\forestregister{folder indent}]!u.parent anchor) |- (.child anchor)}, inner xsep=0.6*#1},
    folder tree indent/.style={before computing xy={l=#1}},
    folder icons/.style={folder, this folder size=#1, folder tree indent=3*#1},
    folder icons/.default={12pt},
}

\newcommand{\lib}[1]{\textsc{#1}}
\newcommand{\obj}[1]{\textit{\texttt{#1}}}
\newcommand{\code}[1]{\lstinline[columns=fixed]{#1}}

\title{\vspace{-2.0cm}\lib{Lingvo}: a Modular and Scalable Framework for Sequence-to-Sequence Modeling\\[+10px]
\normalsize{\url{https://github.com/tensorflow/lingvo}}}
\author{Jonathan~Shen}
\author{Patrick~Nguyen}
\author{Yonghui~Wu}
\author{Zhifeng~Chen}
\author{Mia~X.~Chen}
\author{Ye~Jia}
\author{Anjuli~Kannan}
\author{Tara~Sainath}
\author{Yuan~Cao}
\author{Chung-Cheng~Chiu}
\author{Yanzhang~He}
\author{Jan~Chorowski}
\author{Smit~Hinsu}
\author{Stella~Laurenzo}
\author{James~Qin}
\author{Orhan~Firat}
\author{Wolfgang~Macherey}
\author{Suyog~Gupta}
\author{Ankur~Bapna}
\author{Shuyuan~Zhang}
\author{Ruoming~Pang}
\author{Ron~J.~Weiss}
\author{Rohit~Prabhavalkar}
\author{Qiao~Liang}
\author{Benoit~Jacob}
\author{Bowen~Liang}
\author{HyoukJoong~Lee}
\author{Ciprian~Chelba}
\author{Sébastien~Jean}
\author{Bo~Li}
\author{Melvin~Johnson}
\author{Rohan~Anil}
\author{Rajat~Tibrewal}
\author{Xiaobing~Liu}
\author{Akiko~Eriguchi}
\author{Navdeep~Jaitly}
\author{Naveen~Ari}
\author{Colin~Cherry}
\author{Parisa~Haghani}
\author{Otavio~Good}
\author{Youlong~Cheng}
\author{Raziel~Alvarez}
\author{Isaac~Caswell}
\author{Wei-Ning~Hsu}
\author{Zongheng~Yang}
\author{Kuan-Chieh~Wang}
\author{Ekaterina~Gonina}
\author{Katrin~Tomanek}
\author{Ben~Vanik}
\author{Zelin~Wu}
\author{Llion~Jones}
\author{Mike~Schuster}
\author{Yanping~Huang}
\author{Dehao~Chen}
\author{Kazuki~Irie}
\author{George~Foster}
\author{John~Richardson}
\author{Klaus~Macherey}
\author{Antoine~Bruguier}
\author{Heiga~Zen}
\author{Colin~Raffel}
\author{Shankar~Kumar}
\author{Kanishka~Rao}
\author{David~Rybach}
\author{Matthew~Murray}
\author{Vijayaditya~Peddinti}
\author{Maxim~Krikun}
\author{Michiel~A.~U.~Bacchiani}
\author{Thomas~B.~Jablin}
\author{Rob~Suderman}
\author{Ian~Williams}
\author{Benjamin~Lee}
\author{Deepti~Bhatia}
\author{Justin~Carlson}
\author{Semih~Yavuz}
\author{Yu~Zhang}
\author{Ian~McGraw}
\author{Max~Galkin}
\author{Qi~Ge}
\author{Golan~Pundak}
\author{Chad~Whipkey}
\author{Todd~Wang}
\author{Uri~Alon}
\author{Dmitry~Lepikhin}
\author{Ye~Tian}
\author{Sara~Sabour}
\author{William~Chan}
\author{Shubham~Toshniwal}
\author{Baohua~Liao}
\author{Michael~Nirschl}
\author{Pat~Rondon}
\affil{}
\date{\monthyeardate\today}

\begin{document}

\begin{titlepage}
\footnotetext[2]{Special thanks to Alexander Grushetsky and Adam Sadovsky for the initial design of the Params class.}
\maketitle
\end{titlepage}

\begin{abstract}
 Lingvo is a Tensorflow framework offering a complete solution for collaborative deep learning research, with a particular focus towards sequence-to-sequence models. Lingvo models are composed of modular building blocks that are flexible and easily extensible, and experiment configurations are centralized and highly customizable. Distributed training and quantized inference are supported directly within the framework, and it contains existing implementations of a large number of utilities, helper functions, and the newest research ideas. Lingvo has been used in collaboration by dozens of researchers in more than 20 papers over the last two years. This document outlines the underlying design of Lingvo and serves as an introduction to the various pieces of the framework, while also offering examples of advanced features that showcase the capabilities of the framework.
\end{abstract}

\tableofcontents

\newpage

\section{Introduction}

This paper presents the open-source \lib{Lingvo} framework developed by \lib{Google} for sequence modeling with deep neural networks. 
To date, this framework has produced a number of state-of-the-art results in machine translation~\cite{bestofboth,directseq2seq,wu2016gnmt}, speech recognition~\cite{monotone,Chiu18,callisto,Anjuli18,hardalign,multidialect,asr-compression,PrabhavalkarSainathWu18,nolex,neuraltransducer,e2e-multilingual,contextual}, speech synthesis~\cite{tacotron2v,tacotron2d,tacotron2}, and speech translation~\cite{foreignspeech,jia2018leveraging}. It is currently being used by dozens of researchers in their day-to-day work.

We begin by motivating the design of the framework in Section~\ref{s:design}, including the development environment it was built for as well as its guiding principles. That is followed by an exposition of its core components and the role possessed by each of them.

Then, in Section~\ref{s:implementation}, we take a deeper dive into how fundamental concepts are implemented and what that means for users of the framework. This covers topics such as how trainable variables are managed and how hyperparameters are configured, as well as the basic APIs involved in composing layers into a model. While there will be some code snippets, those seeking complete examples with code should refer to the codelab~\cite{codelab}.

Section~\ref{s:lifeofatrainingrun} provides a consolidated walk-through of the flow of logic during a training run. It outlines the pieces involved from how the model is constructed to how its parameters are updated.

Finally, advanced usage such as distributed training, multi-task models, and inference are described in Section~\ref{s:advancedusage}.

\section{Design}\label{s:design}

\subsection{Motivation}

In research, it is critical to be able to quickly prototype and iterate on new ideas. But, when working in a collaborative environment, it is also critical to be able to easily reuse code and document past experiments.

\lib{Lingvo} evolved out of the need to support a large group of applied researchers working on speech and natural language problems in a single shared codebase.

It follows these guiding principles:
\begin{itemize}
    \item Individual pieces should be small and modular, implement the same consistent interface, and be easily extensible;
    \item Experiments should be shared, comparable, reproducible, understandable, and correct;
    \item Performance should efficiently scale to production-scale datasets and distributed training over hundreds of accelerators; 
    \item Code should be shared as much as possible when transitioning from research to production.
\end{itemize}

\paragraph{Modular building blocks.}

\lib{Lingvo} is designed for collaboration, focusing on code with a consistent interface and style that is easy to read and understand, and a flexible modular layering system that promotes code reuse. The same building blocks, such as LSTM or attention layers, can be used as-is across different models with assurance of good quality and performance. Because the blocks are general, an algorithmic improvement in one task (such as the use of multi-head attention in Machine Translation) can be immediately applied to another task (e.g.\ Speech Recognition). With many people using the same codebase, this makes it extremely easy to employ ideas others are trying in your own models. This also makes it simple to adapt existing models to new datasets.

The building blocks are each individual classes, making it straightforward to extend and override their implementation. Layers are composed in a hierarchical manner, separating low-level implementation details from high-level control flow.

\paragraph{Shared, comparable, reproducible, understandable, and correct experiments.} 

A big problem in research is the difficulty in reproducing and comparing results, even between people working in the same team. To better document experiments and allow the same experiment to be re-run in the future, \lib{Lingvo} adopts a system where all the hyperparameters of a model are configured in their own dedicated sub-directory separate from the model logic and are meant to be committed to a shared version control system. As the models are built from the same common layers, this allows our models to be compared with each other without worrying about effects from minute differences in implementation.

All models follow the same overall structure from input processing to loss computation, and all the layers have the same interface. In addition, all the hyperparameters are explicitly declared and their values are logged at runtime. Finally, there are copious amounts of assertions about tensor values and shapes as well as documentation and unit tests. This makes it very easy to read and understand new models when familiar with the framework, and to debug and ensure that the models are correct.

\paragraph{Performance.} 

\lib{Lingvo} is used to train on production-scale datasets. As a matter of necessity, its implementation has been optimized, from input processing to the individual layers. 
Support for synchronous and asynchronous distributed training is provided.

\paragraph{Deployment-readiness.}

Ideally, there should be little porting from research to product deployment. In \lib{Lingvo}, inference-specific graphs are built from the same shared code used for training, and individual classes can be overwritten with device-specific implementations while the high level model architecture remains the same. In addition, quantization support is built directly into the framework.

However, these benefits come at the cost of more discipline and boilerplate, a common trade-off between scalability and fast prototyping.

\subsection{Components}\label{ss:components}

The following are the core components of the \lib{Lingvo} framework.

\paragraph{Models:}
A \obj{Model} is an abstract collection of one or more \obj{Tasks}. For single-task models the \obj{Model} is just a transparent wrapper around the \obj{Task} and the two can be considered the same. For multi-task models, the \obj{Model} controls how variables are shared between \obj{Tasks} and how \obj{Tasks} are sampled for training.

\paragraph{Tasks:}
A \obj{Task} is a specification of a complete optimization problem, such as image classification or speech recognition. It contains an input generator, \obj{Layers} representing a neural network, a loss value, and an optimizer, and is in charge of updating the model parameters on each training step.

\paragraph{Layers:}
A \obj{Layer} represents an arbitrary function possibly with trainable parameters. A \obj{Layer} can contain other \obj{Layers} as children. \code{SoftMax}, \code{LSTM}, \code{Attention}, and even a \obj{Task} are all examples of \obj{Layers}.

\paragraph{Input Generators:}
\lib{Lingvo} input generators are specialized for sequences, allowing batching input of different lengths in multiple buckets and automatically padding them to the same length. Large datasets that span multiple input files are also supported. The flexibility of the \code{generic_input} function enables simple and efficient implementations of custom input processors.

\paragraph{Params:}
The \obj{Params} object contains hyperparameters for the model. They can be viewed as local versions of \code{tf.flags}. \obj{Layers}, \obj{Tasks}, and \obj{Models} are all constructed in accordance to the specifications in their \obj{Params}.

\obj{Params} are hierarchical, meaning that the \obj{Params} for an object can contain \obj{Params} configuring child objects.

\paragraph{Experiment Configurations:}
Each experiment is defined in its own class and fully defines all aspects of the experiment from hyperparameters like learning rate and optimizer parameters to options that affect the model graph structure to input datasets and other miscellaneous options. 

These standalone configuration classes make it easy to keep track of the params used for each experiment and to reproduce past experiments. It also allows configurations to inherit from other configurations.

All experiment params are registered in a central registry, and can be referenced by its name, e.g.\ \code{image.mnist.LeNet5}.

\paragraph{Job Runners:}
\lib{Lingvo}'s training setup is broken into separate jobs. For example, the \obj{Controller} job is in charge of writing checkpoints while the \obj{Evaler} job evaluates the model on the latest checkpoint. For a full description of the different job runners see Section~\ref{ss:distributedtraining}.

\paragraph{NestedMap:}
A \obj{NestedMap} is a generic dictionary structure for arbitrary structured data similar to \code{tf.contrib.framework.nest}. It is used throughout \lib{Lingvo} to pass data around. Most python objects in the code are instances of either \obj{Tensor}, a subclass of \obj{BaseLayer}, or \obj{NestedMap}.

\paragraph{Custom Ops:}
\lib{Lingvo} supports custom op kernels written in \code{C++} for high-performance code. For example, custom ops are used for the input pipeline, beam search, and tokenization.

\section{Implementation}\label{s:implementation}

This section provides a more detailed look into the core \lib{Lingvo} APIs. Section~\ref{ss:params} introduces the \obj{Params} class which is used to configure everything. Section~\ref{ss:layers} covers how \obj{Layers} are constructed, how they work, and how they can be composed. Section~\ref{ss:variablemanagement} describes how variables are created and managed by each \obj{Layer}. Section~\ref{ss:inputprocessing} goes over input reading and processing, and Sections~\ref{ss:modelregistration}, \ref{ss:overridingparamsfromthecommandline}, and \ref{ss:assertions} briefly go over model registration, overriding params, and runtime assertions. Finally, Section~\ref{ss:codelayout} gives a simple overview of the layout of the source code.

\subsection{Params}\label{ss:params}

The \obj{Params} class is a dictionary with explicitly defined keys used for configuration. Keys should be defined when the object is created, and trying to access or modify a nonexistent key will raise an exception. In practice, every \obj{Layer} has a \code{Params} classmethod, which creates a new params object and defines the keys used to configure the layer with a reasonable default value. Then, in a separate experiment configuration class, these default values are overridden with experiment-specific values.

\subsection{Layers}\label{ss:layers}

In order to construct a \obj{Layer}, an instance of these layer's \obj{Params} is required. The params includes details such as:\begin{itemize}
    \item \code{cls}: the layer's class, 
    \item \code{name}: the layer's name, and
    \item \code{params_init}: how the variables created by this layer should be initialized.
\end{itemize}
Because the class is contained in the params, the following ways of constructing the layer are equivalent:

\begin{lstlisting}
p = SomeLayerClass.Params()
layer = SomeLayerClass(p)  # Call the constructor.
layer = p.cls(p)           # Same, but call through the params.
\end{lstlisting}

All layers have a \code{FProp()} function, which is called during the forward step of a computation. Child layers can be created in the constructor using \code{self.CreateChild('child_name', child_params)}, and they can be referenced by \code{self.child_name}.

\subsection{Variable Management}\label{ss:variablemanagement}

Each \obj{Layer} creates and manages its own variables.

Variables are created in the layer's \code{__init__()} method through a call to \code{self.CreateVariable()}, which registers the variable in \code{self.vars} and the value of the variable (potentially after a transform like adding variational noise) in \code{self.theta}. In \code{FProp()}, because it may be executed on different devices in distributed training, for performance reasons it is best to access the variables through the \code{theta} parameter passed in to the function rather than \code{self.vars} or \code{self.theta}.

Variable placement is determined by the \code{cluster.GetPlacer()} function. The default policy is to place each variable on the parameter server that has the least bytes allocated. For model parallelism, an explicit policy based on e.g.\ variable scope can be adopted.

There are many benefits to explicitly managing variables instead of using \code{tf.get_variable}:
\begin{itemize}
    \item It supports research ideas such as weight noise.
    \item The \code{variable_scope} construct can be error prone and less readable, for example accidental reuse of a variable.
    \item For sync replica training, sharing the weights between multiple workers on the same machine is otherwise awkward.
\end{itemize}

\subsection{Input Processing}\label{ss:inputprocessing}

\lib{Lingvo} supports inputs in either plain text or \code{TFRecord} format. Sequence inputs can be bucketed by length through the \code{bucket_upper_bound} and \\
\code{bucket_batch_limit} params.

A tokenizer can be specified for text inputs. Available tokenizers include \code{VocabFileTokenizer} which uses a look-up table provided as a file, \code{BpeTokenizer} for byte pair encoding~\cite{bpe}, and \code{WpmTokenizer} for word-piece models~\cite{wpm}.

The input file pattern should be specified as ``\code{type:glob_pattern}'' through the \code{file_pattern} param. The input processor should implement the \\
\code{_DataSourceFromFilePattern()} method, which returns an op that when executed reads from the file and returns some tensors. Often this op is implemented as a custom \code{C++} op using the \code{RecordProcessor} interface. The tensors returned by this op can be retrieved by calling \code{_BuildDataSource()}, and can be used to fill in an input batch \obj{NestedMap} to be returned by the \code{InputBatch()} method. Finally, batch-level preprocessing can also be implemented in \code{PreprocessInputBatch()}.

In addition to using a custom \code{RecordProcessor} op, an input processor can also be defined directly in \code{Python} through the \code{generic_input} op.

\subsection{Model Registration}\label{ss:modelregistration}

Configuration classes lie inside \path{lingvo/tasks/<task>/params/<param>.py} and are annotated with \code{@model_registry.RegisterSingleTaskModel} for the typical case of a single-task model. This annotation adds the class to the model registry with a key of \code{<task>.<param>.<classname>} (e.g.\ \code{image.mnist.LeNet5}).

The class should be a subclass of \code{SingleTaskModelParams} and implement the \code{Task()} method, which returns a \obj{Params} instance configuring a \obj{Task}. The registration code will automatically wrap the \obj{Task} into a \code{SingleTaskModel}.

The class should also implement the \code{Train()}, \code{Test()}, and maybe \code{Dev()} methods. These methods return a \obj{Params} instance configuring an input generator, and represent different datasets.

An example is shown in Figure~\ref{fig:registermodel}.

\begin{figure}[ht]
\begin{lstlisting}
@model_registry.RegisterSingleTaskModel
class MyTaskParams(base_model_params.SingleTaskModelParams):
  @classmethod
  def Train(cls):
    ...  # Input params.

  @classmethod
  def Task(cls):
    p = my_model.MyNetwork.Params()
    p.name = 'my_task'
    ...
    return p
\end{lstlisting}
\caption{Registering a single-task model.}\label{fig:registermodel}
\end{figure}

\subsection{Overriding Params from the Command Line}\label{ss:overridingparamsfromthecommandline}

It is possible to override the values of any hyperparameter for a specific run using the \code{--model_params_override} or \code{--model_params_file_override} flags. This makes it simple to start similar jobs for hyperparameter tuning.

\subsection{Assertions}\label{ss:assertions}

\path{py_utils.py} contains functions for run-time assertions about values and shapes as well as \code{CheckNumerics()} for detecting NaNs. Assertions can be disabled with the command-line \code{--enable_asserts=false}. Similarly, \code{CheckNumerics} can be disabled with \code{--enable_check_numerics=false}.

\subsection{Code Layout}\label{ss:codelayout}

\begin{forest}
    for tree={font=\sffamily, grow'=0,
    folder indent=.9em, folder icons,
    edge=dotted, align=left}
    [lingvo
      [{trainer.py\\[-1ex]\scriptsize Entry point.}, is file]
      [{model\_imports.py\\[-1ex]\scriptsize Imports and registers all model params in the global registry.}, is file]
      [core
          [base\_input\_generator.py, is file]
          [base\_layer.py, is file]
          [base\_model.py, is file]
          [{cluster.py\\[-1ex]\scriptsize Contains the policy for op placement.}, is file]
          [hyperparams.py, is file]
          [{attention.py, layers.py, rnn\_cell.py, rnn\_layers.py\\[-1ex]\scriptsize Contains implementations for many common layers.}, is file]
          [optimizer.py, is file]
          [{py\_utils.py\\[-1ex]\scriptsize Most utility functions are here.}, is file]
          [{recurrent.py\\[-1ex]\scriptsize The functional RNN.}, is file]
          [{summary\_utils.py\\[-1ex]\scriptsize Contains utilities for dealing with summaries.}, is file]
          [{ops\\[-1ex]\scriptsize Folder for custom \code{C++} ops.}
            [{record\_*.*\\[-1ex]\scriptsize The input processing pipeline.}, is file]
            [{py\_x\_ops.py\\[-1ex]\scriptsize Python bindings for the \code{C++} ops.}, is file]
            [{x\_ops.cc\\[-1ex]\scriptsize \code{C++} op definitions.}, is file]
          ]
      ]
      [tasks
          [{<task>\\[-1ex]\scriptsize Folder for an individual task/domain/project.}
            [{params\\[-1ex]\scriptsize Folder for model params.}]
          ]
      ]
      [tools]
    ]
\end{forest}

\section{Life of a Training Run}\label{s:lifeofatrainingrun}

\begin{figure}[h]
\centering
\includegraphics[width=\textwidth]{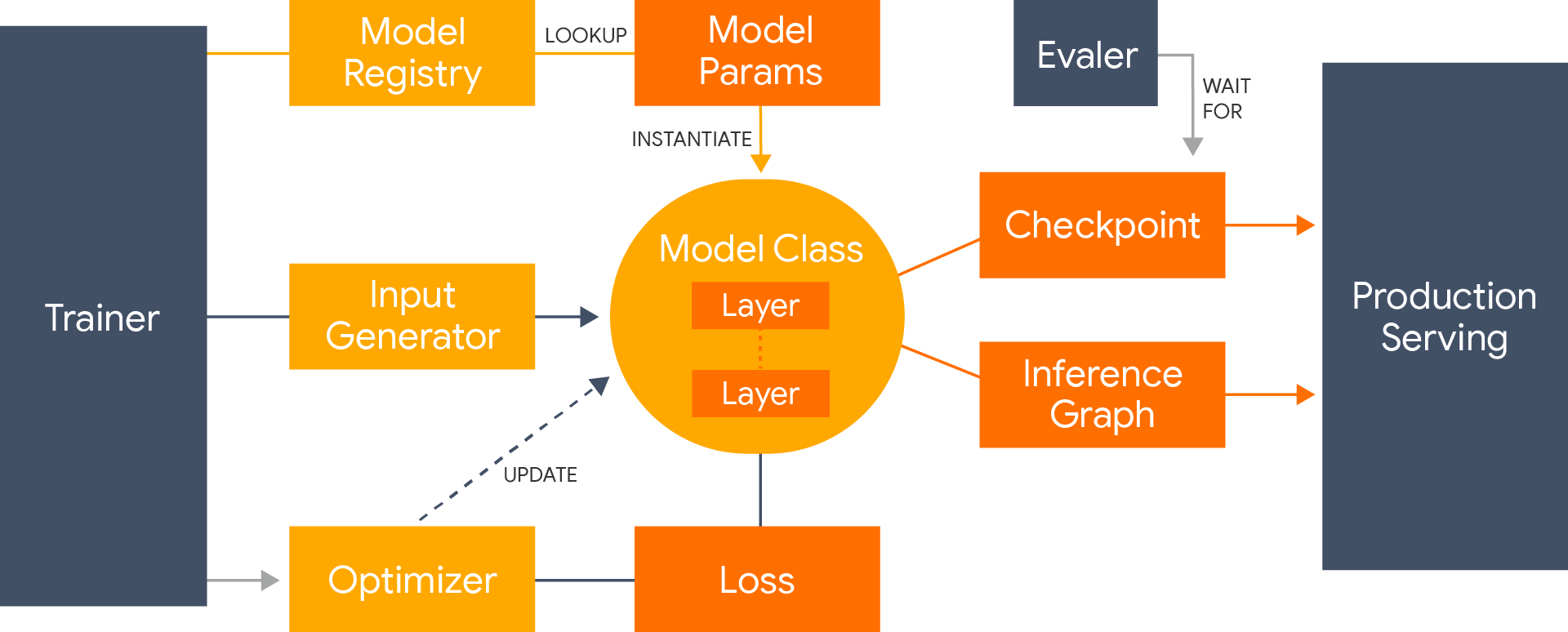}
\caption{An overview of the \lib{Lingvo} framework, outlining how models are instantiated, trained, and exported for evaluation and serving.}
\end{figure}

This section gives an overview of what happens behind the scenes from the start of a training run to the end of the first step for a single-task model.

Training is started by launching the \code{trainer} with the name of a model and the path to a log directory. The model name is resolved using the model registry to obtain the \obj{Params} for the model, and the various job runners for training are created.

The \obj{Params} at this point is just the params for the top level \obj{Model} with the overrides in the experiment configuration corresponding to the model name specified. No \obj{Layers} have been instantiated yet.

Each runner then independently instantiates the model and builds the tensorflow graphs and ops that they will execute based on their job. For example, the \obj{Trainer} will build a \code{train_op} spanning both the \code{FProp()} and \code{BProp()} graphs and involves updating model parameters, while the \obj{Evaler} and \obj{Decoder} will build a \code{eval_metrics} op involving only the \code{FProp()} graph with \code{p.is_eval} set. There can be multiple \obj{Evalers} and \obj{Decoders}, one for each evaluation dataset.

Instantiating the model calls its \code{__init__()} method, which constructs the \obj{Params} for child layers and instantiates them recursively through calls to \code{self.CreateChild()}. These child layer params could be exposed as part of the top level model params, perhaps as a ``params template'', allowing them to be configured in the params files, or they could be constructed completely from scratch in the \code{__init__()} method based on the information available at that time. The \code{__init__()} method is also in charge of creating the variables managed by the \obj{Layer} through \code{self.CreateVariable()}.

Once the graphs are built, the \obj{Trainer} runner will wait for the \obj{Controller} runner to initialize the variables or restore them from a checkpoint, while the evaluation runners will wait for a new checkpoint to be written to the log directory.

After the variables are initialized, the \obj{Trainer} will run training, i.e.\ calling \code{session.run} with the model's \code{train_op}, in a loop, and the \obj{Controller} will produce training summaries for each step. When enough steps have passed, the \obj{Controller} writes a new checkpoint to the log directory. The \obj{Evaler} and \obj{Decoder} detect this new checkpoint and evaluates the model at that checkpoint to generate summaries. This process then continues until the user terminates all jobs or \code{p.train.max_steps} is reached.

For more details about the various runners such as the difference between Evalers and Decoders, as well as information about which devices individual ops in the graph will be placed on during distributed training, see Section~\ref{ss:distributedtraining}.

\section{Advanced Usage}\label{s:advancedusage}

This section provides some examples of advanced features. This is by no means an exhaustive list of all the existing features, and many new features are continually being added.

Section~\ref{ss:distributedtraining} describes the distributed training setup. 
Section~\ref{ss:multitaskmodels} details how multi-task models are configured and registered, and Section~\ref{ss:inferenceandquantization} gives a brief look into inference and productionization support.

\subsection{Distributed Training}\label{ss:distributedtraining}

Both synchronous as well as asynchronous distributed training are supported. In asynchronous mode, each individual worker job executes its own training loop and is completely independent from the other workers. In synchronous mode, there is a single training loop driven by a trainer client that distributes work onto the various worker jobs.

Here we summarize the different types of job runners under each configuration.

A shared directory on a distributed file system where checkpoints can be written and loaded is assumed to exist.

\subsubsection{Common Runners}

\paragraph{Controller:} This job handles initializing variables and saving/loading checkpoints as well as writing training summaries.
\paragraph{Evaler:} This job loads the latest checkpoint and runs and exports evaluation summaries. Multiple evalers can be started for different datasets.
\paragraph{Decoder:} This job loads the latest checkpoint and runs and exports decoding summaries. Multiple decoders can be started for different datasets. Decoders are different from evalers in that the ground-truth is used during evaluation but not during decoding. A concrete example is that Evalers can use teacher-forcing while Decoders may need to rely on beam search.

\subsubsection{Asynchronous Training}

\paragraph{Trainer:} This is the worker job which runs the training op and sends variable updates.
\paragraph{Parameter Server:} Variable values are stored here. Trainer jobs send updates and receive global values periodically.
\paragraph{Data Processor:} This is an optional job for loading data before dispatching them to trainers, to offload the cost associated with loading and preprocessing data from the trainer to a separate machine.

\subsubsection{Synchronous Training}
\paragraph{Worker:} The worker job in sync training runs the training op like the trainer job in async training but they do not perform variable updates.
\paragraph{Trainer client:} The trainer client drives the training loop and aggregates their results before updating the variables. There are no parameter servers in sync training. Instead, the worker jobs act as parameter servers, and the trainer client sends the relevant variable updates to each worker.

\subsection{Multi-task Models}\label{ss:multitaskmodels}

A multi-task model is composed of individual \obj{Tasks} sharing variables. Existing options for variable sharing range from sharing just the encoder\\
(\code{multitask_model.SharedEncoderModel}) to fine-grained control with\\
\code{multitask_model.RegExSharedVariableModel}.

Multi-task model params should be a subclass of \code{MultiTaskModelParams} and implement the \code{Model()} method, which returns a \obj{Params} instance configuring a \code{MultiTaskModel}. The \code{task_params} and \code{task_probs} attributes define respectively the params and relative weight of each \obj{Task}.

An example of registering a multi-task model is shown in Figure~\ref{fig:registermultitask}.

\begin{figure}[h!]
\begin{lstlisting}
@model_registry.RegisterMultiTaskModel
class MyMultiTaskParams(base_model_params.MultiTaskModelParams):
  @classmethod
  def Train(cls):
    p = super(MyMultiTaskParams, cls).Train()
    task1_input_params = ...
    p.Define('task1', task1_input_params, '')
    # Or, refer to existing single task model params.
    p.Define('task2', MyTaskParams.Train(), '')
    return p

  @classmethod
  def Model(cls):
    p1 = my_model.MyNetwork.Params()
    p1.name = 'task1'
    ...
    # Or, refer to existing single task model.
    p2 = MyTaskParams.Task()
    p2.name = 'task2'

    p = base_model.MultiTaskModel.Params()
    p.name = 'my_multitask_model'
    p.task_params = hyperparams.Params()
    p.task_params.Define('task1', p1, '')
    p.task_params.Define('task2', p2, '')
    p.task_probs = hyperparams.Params()
    p.task_probs.Define('task1', 0.5, '')
    p.task_probs.Define('task2', 0.5, '')
    return p
\end{lstlisting}
\caption{Registering a multi-task model.}
\label{fig:registermultitask}
\end{figure}

Knowledge distillation is also supported via \code{base_model.DistillationTask}. For knowledge distillation, the teacher parameters must be loaded from a checkpoint file by specifying \code{params.train.init_from_checkpoint_rules} in the \code{Task()} definition.

\subsection{Inference and Quantization}\label{ss:inferenceandquantization} 
Once models have been trained, they must be deployed on a server or on an embedded device. Typically, during inference, models will be executed on a device with fixed-point arithmetic. To achieve the best quality, the dynamic range must be kept in check during training. We offer quantized layers that wrap the training and inference computation functions for convenience.

Inference has different computational characteristics than training. For latency reasons, the batch size is smaller, sometimes even equal to just 1. For sequence models, often a beam search is performed. It may even be preferable to drive inference one timestep at a time. Several constraints dominate the choice of how to run the computation: 1) available operations, 2) desired latency, 3) parallelizability, and 4) memory and power consumption. To enable the greatest amount of flexibility given these constraints, we leave it to the designer of the model to express inference in the optimal way by explicitly exporting inference graphs rather than leaving it to a graph converter. A basic inference graph can be written in a few lines of code, reusing the same functions used for building the training graph, while for more complicated inference graphs it is possible to even completely swap out the implementation of a low level layer.

\clearpage
\bibliographystyle{plain}
\bibliography{lingvo}

\end{document}